**Vasile PĂTRAŞCU, Vasile BUZULOIU**
Image Processing and Analysis Laboratory (LAPI),
"Politehnica" University of Bucharest,
Bd. Iuliu Maniu 1-3, Bucharest, Romania
`vpatrascu@hotmail.com,vbuzuloiu@alpha.imag.pub.ro`

# THE AFFINE TRANSFORMS FOR IMAGE ENHANCEMENT IN THE CONTEXT OF LOGARITHMIC MODELS

**Abstract**.
The logarithmic model offers new tools for image processing. An efficient method for image enhancement is to use an affine transformation with the logarithmic operations: addition and scalar multiplication. We define some criteria for automatically determining the parameters of the processing and this is done via mean and variance computed by logarithmic operations.

## 1　INTRODUCTION

Image enhancement is a pre-processing step applied to any input image of an image analysis system. The enhancement methods are specific for every application and the physical nature of pictures as well as the mathematical models for their representation are equally important for choosing the right enhancement method. For the images obtained by reflected light, where the effects are naturally of a multiplicative form, Stockham [5] proposed an enhancement method based on the homomorphic theory introduced by Oppenheim [2]. The key to this approach is a homomorphism, which transforms the product into a sum (by logarithm) and so allows one to use the classical linear filtering. Another logarithmic representation evolving from multiplicative properties of the transmission of light was developed by Jourlin and Pinoli [1]. Their model was more elaborate from the mathematical point of view. Still, both mentioned models do work with semi-bounded sets of values. Our logarithmic model [3,4] does work with bounded real sets: the range of values of our images, say $[0, M]$, is linearly applied onto the standard set $(-1,1)$ and it is this interval that plays the central role in the model: it is endowed with the structure of a linear (moreover: Euclidean) space over the scalar field of real numbers $R$. In this framework we can obtain very efficient enhancement algorithms with just one affine transformation.

## 2  THE FUNDAMENTALS OF THE LOGARITHMIC MODEL

The most usual mathematical model for the gray level images is the real valued function defined on a bounded subset of $R^2$. Having in mind that the function value in a point $(x, y)$ representing the luminosity of that pixel or reflectancy or transparency - it becomes clear that the functions we use are bounded ones (say, they take values in a bounded interval $[0, M)$). The problems appear when processing an image: the mathematical operations on real valued functions use implicitly the algebra of the real numbers i.e. on the whole real axis and we are faced with results that do not belong anymore to the interval $[0, M]$ - the only ones with physical meaning. Nevertheless this situation is generally accepted by using a truncation of the values out of the range $[0, M]$. The framework used in this paper, from this point of view, is radically different: namely, we want our set $[0, M]$ to be a stable set with respect to the algebraic operations that we use - addition and scalar multiplication (with real scalars). For convenience we prefer to build this structure on the set $(-1,1)$ and move $[0, M]$ onto it by a linear transformation. In the set of gray levels $E = (-1,1)$ we will define the addition $\langle + \rangle$ and the multiplication $\langle \times \rangle$ by a real scalar and then, defining a scalar product $(\cdot | \cdot)_E$ and a norm $\| \cdot \|_E$, we shall obtain an Euclidean space of gray levels.

### 2.1  The addition

We shall define the sum of two gray levels, $v_1 \langle + \rangle v_2$, by:

$$v_1 \langle + \rangle v_2 = \frac{v_1 + v_2}{1 + v_1 v_2}, \forall v_1, v_2 \in E \qquad (1)$$

The neutral element for the addition is $\theta = 0$ and each element $v \in E$ has an opposite $w = -v$. The addition $\langle + \rangle$ is stable, associative, and commutative. Thus, it follows that this operation induces on $E$ a commutative group. We shall define the subtraction operation $\langle - \rangle$ by:

$$v_1 \langle - \rangle v_2 = \frac{v_1 - v_2}{1 - v_1 v_2}, \forall v_1, v_2 \in E \qquad (2)$$

Using the defined subtraction $\langle - \rangle$, we shall denote the opposite of $v$ by $\langle - \rangle v$.

### 2.2  The multiplication by a scalar

We shall define the multiplication $\langle \times \rangle$ of a gray level $v$ by a scalar $\lambda$ as:

$$\lambda \langle \times \rangle v = \frac{(1+v)^\lambda - (1-v)^\lambda}{(1+v)^\lambda + (1-v)^\lambda}, \forall v \in E, \forall \lambda \in R \qquad (3)$$

The operations defined above, the addition $\langle + \rangle$ and the scalar multiplication $\langle \times \rangle$ induce on $E$ a real vector space structure.

## 2.3 The fundamental isomorphism

The vector space of gray levels $(E, \langle + \rangle, \langle \times \rangle)$ is isomorphic to the space of real numbers $(R, +, \cdot)$ by the function $\varphi : E \to R$, defined as:

$$\varphi(v) = \frac{1}{2} \ln\left(\frac{1+v}{1-v}\right), \forall v \in E \qquad (4)$$

The isomorphism $\varphi$ verifies:

$$\varphi(v_1 \langle + \rangle v_2) = \varphi(v_1) + \varphi(v_2), \forall v_1, v_2 \in E \qquad (5)$$
$$\varphi(\lambda \langle \times \rangle v) = \lambda \cdot \varphi(v), \forall \lambda \in R, \forall v \in E \qquad (6)$$

The particular nature of this isomorphism induces the logarithmic character of the mathematical model.

## 2.4 The Euclidean space of gray levels

The scalar product of two gray levels, $(\cdot | \cdot)_E : E \times E \to R$ is defined with respect to the isomorphism from (4) as:

$$(v_1 | v_2)_E = \varphi(v_1) \cdot \varphi(v_2), \forall v_1, v_2 \in E \qquad (7)$$

Based on the scalar product $(\cdot | \cdot)_E$ the gray level space becomes an Euclidean space. The norm $\| \cdot \|_E : E \to R^+$ is defined via the scalar product:

$$\| v \|_E = \sqrt{(v, v)_E} = |\varphi(v)|, \forall v \in E \qquad (8)$$

## 3  GRAY LEVEL IMAGE ENHANCEMENT BY AFFINE TRANSFORMS

A gray level image is a function defined over the spatial domain $D \subset R^2$ and takes values within the gray level set $E$. We shall denote by $F(D, E)$ the set of gray level images defined over $D$. We shall thus consider the class of affine transforms on $F(D, E)$, that is, the transforms, $\psi : F(D, E) \to F(D, E)$, of the form: $\psi(f) = \alpha \langle \times \rangle (f \langle + \rangle \beta)$. Our approach consists of choosing $\alpha$ and $\beta$ such that the resulting image has the same average gray level and gray level variance as a uniformly distributed random variable within the set $E = (-1, 1)$ (i.e. $\mu_u = 0$, $\sigma_u^2 = 1/3$). For any given image $f$, with mean $\mu_f$ and variance $\sigma_f^2$, the affine transform becomes: $\psi(f) = \frac{\sigma_u}{\sigma_f} \langle \times \rangle (f \langle - \rangle \mu_f)$. In the discrete case $\mu_f$ and $\sigma_f^2$ are defined by: $\mu_f = \underset{(x,y) \in D}{\langle + \rangle} \left( \frac{1}{card(D)} \langle \times \rangle f(x,y) \right)$ and $\sigma_f^2 = \sum_{(x,y) \in D} \frac{1}{card(D)} \cdot \| f(x,y) \langle - \rangle \mu_f \|_E^2$ where $card(D)$ is the cardinality of the support $D$.

**4   EXPERIMENTAL RESULT**

In order to get some practical results, the proposal method was used to enhance some dark, bright or low contrast images. To exemplify, four images were picked out: one dark ("news") in Fig.1a, one bright ("cells") in Fig.2a, two with low contrast ("lax" and "miss") in Fig.3a respectively Fig.4a. The images were enhanced with the following affine transforms: $\psi_1(v) = 2.37 \langle \times \rangle (v \langle + \rangle 0.71)$ for "news", $\psi_2(v) = 3.02 \langle \times \rangle (v \langle - \rangle 0.65)$ for "cells", $\psi_3(v) = 4.69 \langle \times \rangle (v \langle + \rangle 0.01)$ for "lax" and $\psi_4(v) = 2.03 \langle \times \rangle (v \langle - \rangle 0.08)$ for "miss".

The graphics of the affine transforms are shown in Fig.1b, Fig.2b, Fig.3b and Fig.4b. The enhanced images can be seen in Fig.1c, Fig.2c, Fig.3c and Fig.4c.

**5   CONCLUSIONS**

We have presented a mathematical model for the gray level images by defining an algebraic structure on a bounded interval, and by introducing some basic operations (addition, scalar multiplication) and functions (scalar product, norm). This structure, based on a logarithmic model, provides gray level operations, which yield results that are always confined to the underlying bounded interval of allowed values, and thus avoiding the need of clipping operations. We propose a fully automatic image enhancement method based on the use of an affine transform. The tests show that the proposed techniques allow the automatic correction of the illumination problems that occur during the acquisition process. We claim that the proposed method is just another strong argument for the rich potential of the logarithmic image processing models.

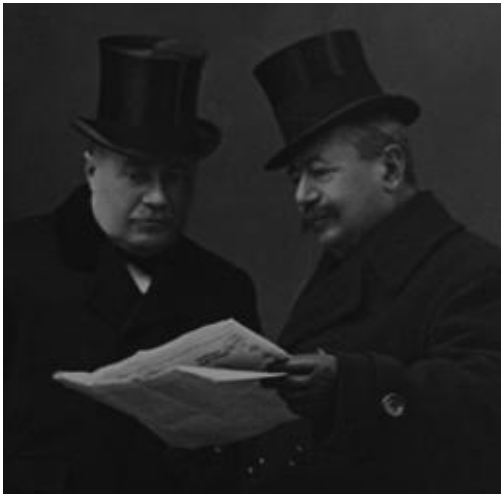

**Fig. 1 a) The original image "news"**

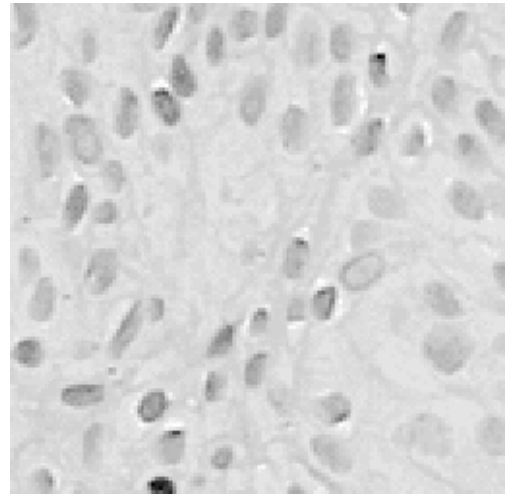

**Fig. 2 a) The original image "cells"**

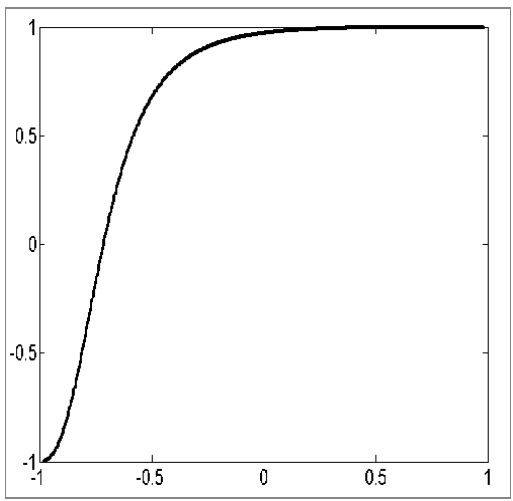

**Fig. 1 b) The log-affine transform for "news"**

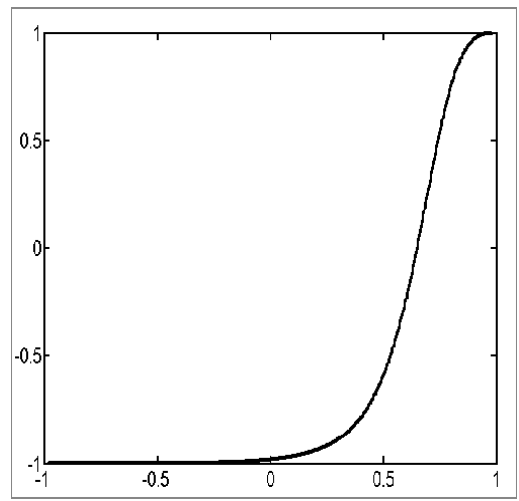

**Fig. 2 b) The log-affine transform for "cells"**

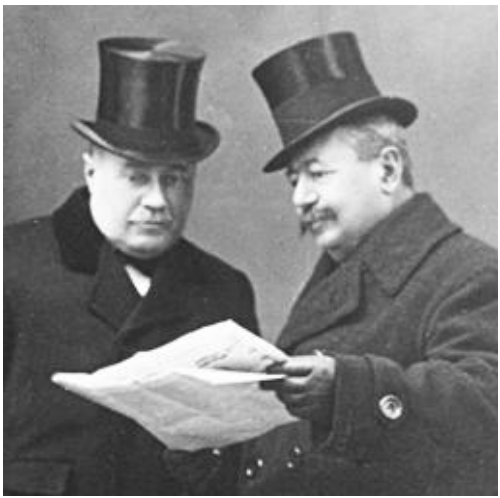

**Fig. 1 c) The log-enhanced image "news"**

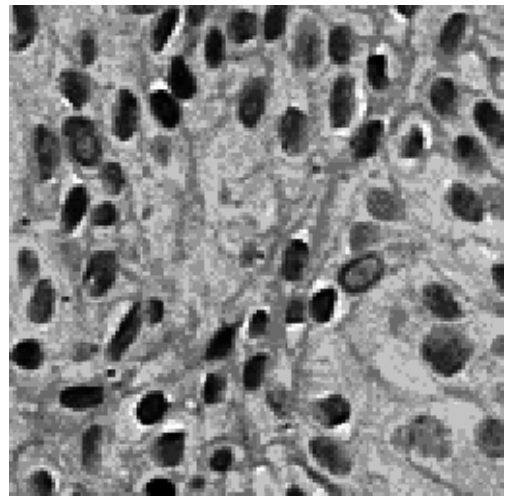

**Fig. 2 c) The log-enhanced image "cells"**

*The Affine Transforms for Image Enhancement in the Context of Logarithmic Models*

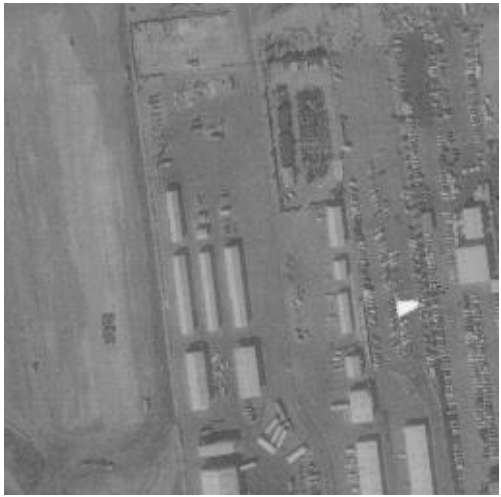

**Fig. 3 a) The original image "lax"**

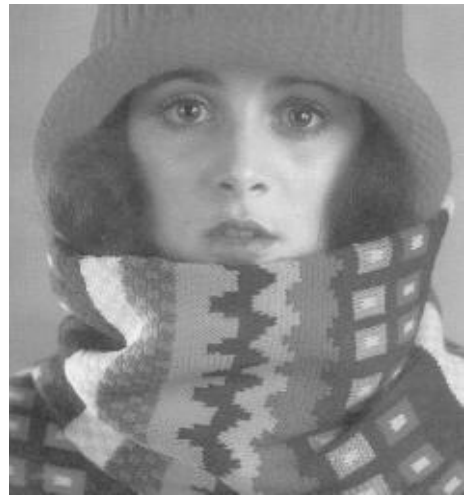

**Fig. 4 a) The original image "miss"**

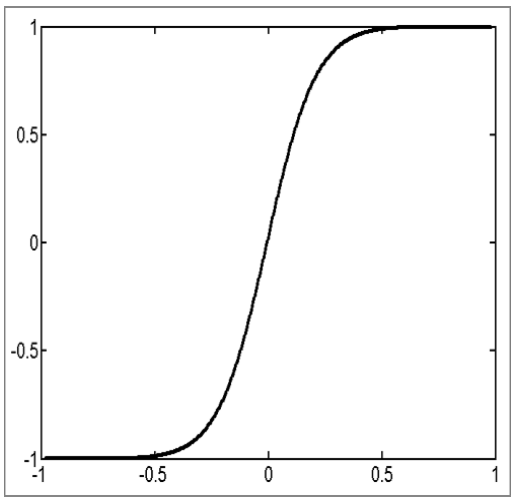

**Fig. 3 b) The log-affine transform for "lax"**

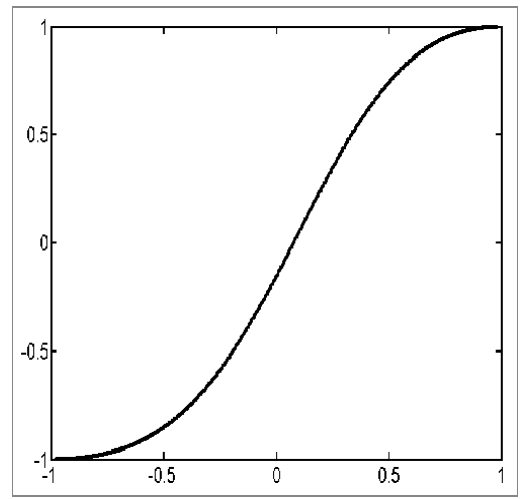

**Fig. 4 b) The log-affine transform for "miss"**

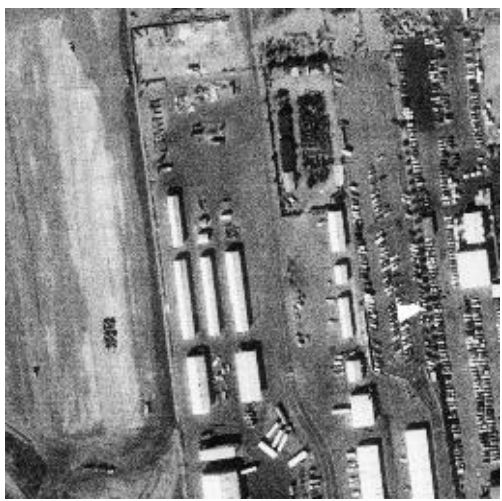

**Fig. 3 c) The log-enhanced image "lax"**

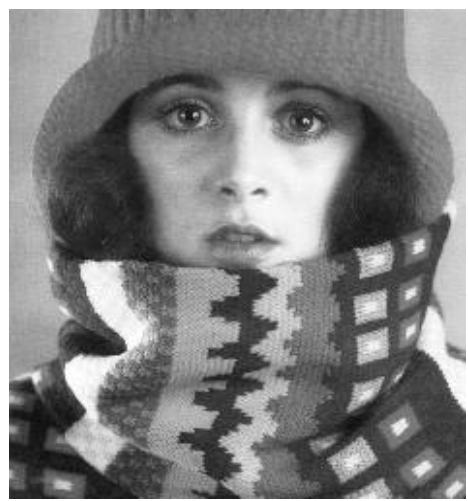

**Fig. 4 c) The log-enhanced image "miss"**